# Open3D: A Modern Library for 3D Data Processing

Qian-Yi Zhou    Jaesik Park    Vladlen Koltun

Intel Labs

## Abstract

*Open3D is an open-source library that supports rapid development of software that deals with 3D data. The Open3D frontend exposes a set of carefully selected data structures and algorithms in both C++ and Python. The backend is highly optimized and is set up for parallelization. Open3D was developed from a clean slate with a small and carefully considered set of dependencies. It can be set up on different platforms and compiled from source with minimal effort. The code is clean, consistently styled, and maintained via a clear code review mechanism. Open3D has been used in a number of published research projects and is actively deployed in the cloud. We welcome contributions from the open-source community.*

## 1. Introduction

The world is three-dimensional. Systems that operate in the physical world or deal with its simulation must often process three-dimensional data. Such data takes the form of point clouds, meshes, and other representations. Data in this form is produced by sensors such as LiDAR and depth cameras, and by software systems that support 3D reconstruction and modeling.

Despite the central role of 3D data in fields such as robotics and computer graphics, writing software that processes such data is quite laborious in comparison to other data types. For example, an image can be efficiently loaded and visualized with a few lines of OpenCV code [3]. A similarly easy to use software framework for 3D data has not emerged. A notable prior effort is the Point Cloud Library (PCL) [18]. Unfortunately, after an initial influx of open-source contributions, PCL became encumbered by bloat and is now largely dormant. Other open-source efforts include MeshLab [6], which provides a graphical user interface for processing meshes; libigl [11], which supports discrete differential geometry and related research; and a variety of solutions for image-based reconstruction [13]. Nevertheless, there is currently no open-source library that is fast, easy to use, supports common 3D data processing workflows, and is developed in accordance with modern software engineering practices.

Open3D was created to address this need. It is an open-source library that supports rapid development of software that deals with 3D data. The Open3D frontend exposes a set of carefully selected data structures and algorithms in both C++ and Python. The Open3D backend is implemented in C++11, is highly optimized, and is set up for OpenMP parallelization. Open3D was developed from a clean slate with a small and carefully considered set of dependencies. It can be set up on different platforms and compiled from source with minimal effort. The code is clean, consistently styled, and maintained via a clear code review mechanism.

Open3D has been in development since 2015 and has been used in a number of published research projects [22, 13, 15, 12]. It has been deployed and is currently running in the Tanks and Temples evaluation server [13].

Open3D is released open-source under the permissive MIT license and is available at http://www.open3d.org. We welcome contributions from the open-source community.

## 2. Design

Two primary design principles of Open3D are usefulness and ease of use [8]. Our key decisions can be traced to these two principles. Usefulness motivated support for popular representations, algorithms, and platforms. Ease of use served as a countervailing force that guarded against heavyweight dependencies and feature creep.

Open3D provides data structures for three kinds of representations: point clouds, meshes, and RGB-D images. For each representation, we have implemented a complete set of basic processing algorithms such as I/O, sampling, visualization, and data conversion. In addition, we have implemented a collection of widely used algorithms, such as normal estimation [16], ICP registration [2, 4], and volumetric integration [7]. We have verified that the functionality of Open3D is sufficient by using it to implement complete workflows such as large-scale scene reconstruction [5, 15].

We carefully chose a small set of lightweight dependencies, including Eigen for linear algebra [9], GLFW for OpenGL window support, and FLANN for fast nearest neighbor search [14]. For easy compilation, powerful but heavyweight libraries such as Boost and Ceres are excluded. Instead we use either lightweight alternatives (e.g., pybind11 instead of Boost.Python) or in-house implementations (e.g., for Gauss-Newton and Levenberg-Marquardt graph optimization). The source code of all dependencies is distributed as part of Open3D. The dependencies can thus be compiled from source if not automatically detected by the configuration script. This is particularly useful for compilation on operating systems that lack package management software, such as Microsoft Windows.

The development of Open3D started from a clean slate and the library is kept as simple as possible. Only algorithms that solve a problem of broad interest are added. If a problem has multiple solutions, we choose one that the community considers standard. A new algorithm is added only if its implementation demonstrates significantly stronger results on a well-known benchmark.

Open3D is written in standard C++11 and uses CMake to support common C++ toolchains including

- GCC 4.8 and later on Linux
- XCode 8.0 and later on OS X
- Visual Studio 2015 and later on Windows

A key feature of Open3D is ubiquitous Python binding. We follow the practice of successful computer vision and deep learning libraries [1, 3]: the backend is implemented in C++ and is exposed through frontend interfaces in Python. Developers use Python as a glue language to assemble components implemented in the backend.

Figure 1 shows code snippets for a simple 3D data processing workflow. An implementation using the Open3D Python interface is compared to an implementation using the Open3D C++ interface and to an implementation based on PCL [18]. The implementation using the Open3D Python interface is approximately half the length of the implementation using the Open3D C++ interface, and about five times shorter than the implementation based on PCL. As an added benefit, the Python code can be edited and debugged interactively in a Jupyter Notebook.

## 3. Functionality

Open3D has nine modules, listed in Table 1.

### 3.1. Data

The *Geometry* module implements three geometric representations: *PointCloud*, *TriangleMesh*, and *Image*.

The *PointCloud* data structure has three data fields: *PointCloud.points*, *PointCloud.normals*, *PointCloud.colors*. They are used to store coordinates, normals, and colors. The master data field is *PointCloud.points*. The other two fields are considered valid only when they have the same number of records as *PointCloud.points*. Open3D provides direct memory access to these data fields via a numpy array. The following code sample demonstrates reading and accessing the point coordinates of a point cloud.

```
from py3d import *
import numpy as np
pointcloud = read_point_cloud('pointcloud.ply')
print(np.asarray(pointcloud.points))
```

Similarly, *TriangleMesh* has two master data fields – *TriangleMesh.vertices* and *TriangleMesh.triangles* – as well as three auxiliary data fields: *TriangleMesh.vertex_normals*, *TriangleMesh.vertex_colors*, and *TriangleMesh.triangle_normals*.

The *Image* data structure is implemented as a 2D or 3D array and can be directly converted to a numpy array. A pair of depth and color images of the same resolution can be combined into a data structure named *RGBDImage*. Since there is no standard depth image format, we have implemented depth image support for multiple datasets including NYU [19], TUM [20], SUN3D [21], and Redwood [5]. The following code sample reads a pair of RGB-D images from the TUM dataset and converts them to a point cloud.

```
from py3d import *
import numpy as np
depth = read_image('TUM_depth.png')
color = read_image('TUM_color.jpg')
rgbd = create_rgbd_image_from_tum_format(color,
    depth)
pointcloud = create_point_cloud_from_rgbd_image(
    rgbd, PinholeCameraIntrinsic.
    prime_sense_default)
```

### 3.2. Visualization

Open3D provides a function *draw_geometries()* for visualization. It takes a list of geometries as input, creates a window, and renders them simultaneously using OpenGL. We have implemented many functions in the visualizer, such as rotation, translation, and scaling via mouse operations, changing rendering style, and screen capture. A code sample using *draw_geometries()* and its result are shown in Figure 2.

In addition to *draw_geometries()*, Open3D has a number of sibling functions with more advanced functionality. *draw_geometries_with_custom_animation()* allows the programmer to define a custom view trajectory and play an animation in the GUI. *draw_geometries_with_animation_callback()* and *draw_geometries_with_key_callback()* accept Python callback functions as input. The callback function is called in an automatic animation loop, or upon a key press event. The following code sample shows a rotating point cloud using *draw_geometries_with_animation_callback()*.

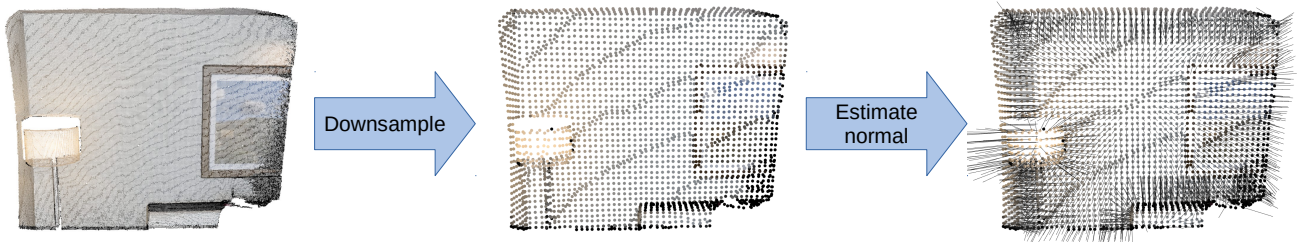

(a) A simple 3D data processing task: load a point cloud, downsample it, and estimate normals.

```
from py3d import *
pointcloud = read_point_cloud('pointcloud.ply')
downsampled = voxel_down_sample(pointcloud, voxel_size = 0.05)
estimate_normals(downsampled, KDTreeSearchParamHybrid(radius = 0.1, max_nn = 30))
draw_geometries([downsampled])
```

(b) An implementation using the Open3D Python interface.

```
#include <Core/Core.h>
#include <IO/IO.h>
#include <Visualization/Visualization.h>
void main(int argc, char *argv[])
{
    using namespace three;
    auto pointcloud = CreatePointCloudFromFile(argv[1]);
    auto downsampled = VoxelDownSample(pointcloud, 0.05)
    EstimateNormals(*downsampled, KDTreeSearchParamHybrid(0.1, 30));
    DrawGeometry(downsampled);
}
```

(c) An implementation using the Open3D C++ interface.

```
#include <pcl/point_types.h>
#include <pcl/point_cloud.h>
#include <pcl/features/normal_3d.h>
#include <pcl/filters/voxel_grid.h>
#include <pcl/io/pcd_io.h>
#include <pcl/visualization/cloud_viewer.h>
void main(int argc, char *argv[])
{
    using namespace pcl;
    PointCloud<PointNormal>::Ptr pcd(new PointCloud<PointNormal>());
    io::loadPCDFile<PointNormal>(argv[1], *pcd);
    VoxelGrid<PointNormal> grid;
    grid.setLeafSize(0.05, 0.05, 0.05);
    grid.setInputCloud(pcd);
    grid.filter(*pcd);
    PointCloud<PointNormal>::Ptr result(new PointCloud<PointNormal>());
    NormalEstimation<PointNormal, PointNormal> est;
    est.setInputCloud(pcd);
    est.setRadiusSearch(0.1);
    est.compute(*result);
    pcl::visualization::PCLVisualizer viewer("PCL Viewer");
    viewer.addPointCloudNormals<PointNormal, PointNormal>(pcd, result, 1, 0.05);
    while (!viewer.wasStopped())
    {
        viewer.spinOnce();
    }
}
```

(d) An implementation based on PCL.

Figure 1: Code snippets for a simple 3D data processing workflow. (a) An illustration of the task. (b) An implementation using the Open3D Python interface. (b) An implementation using the Open3D C++ interface. (c) An implementation based on PCL [18]. The implementation using the Open3D Python interface is dramatically shorter and clearer than the implementation based on PCL.

| Module | Functionality |
|---|---|
| Geometry | Data structures and basic processing algorithms |
| Camera | Camera model and camera trajectory |
| Odometry | Tracking and alignment of RGB-D images |
| Registration | Global and local registration |
| Integration | Volumetric integration |
| I/O | Reading and writing 3D data files |
| Visualization | A customizable GUI for rendering 3D data with OpenGL |
| Utility | Helper functions such as console output, file system, and Eigen wrappers |
| Python | Open3D Python binding and tutorials |

Table 1: Open3D modules.

```
from py3d import *
pointcloud = read_point_cloud('pointcloud.pcd')
mesh = read_triangle_mesh('mesh.ply')
mesh.computer_vertex_normals()
draw_geometries([pointcloud, mesh])
```

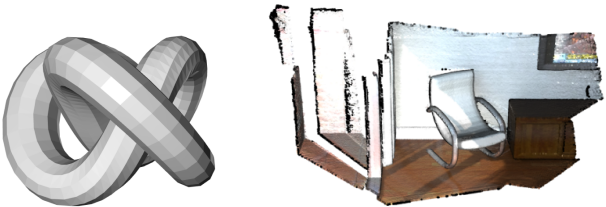

Figure 2: Visualize a mesh and a point cloud using *draw_geometries()*.

```
from py3d import *
pointcloud = read_point_cloud('pointcloud.pcd')
def rotate_view(vis):
  # Rotate the view by 10 degrees
  ctr = vis.get_view_control()
  ctr.rotate(10.0, 0.0)
  return False
draw_geometries_with_animation_callback([
    pointcloud], rotate_view)
```

In the backend of Open3D, these functions are implemented using the *Visualizer* class. This class is also exposed in the Python interface.

### 3.3. Registration

Open3D provides implementations of multiple state-of-the-art surface registration methods, including pairwise global registration, pairwise local refinement, and multiway registration using pose graph optimization. This section gives an example of a complete pairwise registration workflow for point clouds. The workflow begins by reading raw point clouds, downsampling them, and estimating normals:

```
from py3d import *
source = read_point_cloud('source.pcd')
target = read_point_cloud('target.pcd')
source_down = voxel_down_sample(source, 0.05)
target_down = voxel_down_sample(target, 0.05)
estimate_normals(source_down,
    KDTreeSearchParamHybrid(radius = 0.1, max_nn
    = 30))
estimate_normals(target_down,
    KDTreeSearchParamHybrid(radius = 0.1, max_nn
    = 30))
```

We then compute FPFH features and apply a RANSAC-based global registration algorithm [17]:

```
source_fpfh = compute_fpfh_feature(source_down,
    KDTreeSearchParamHybrid(radius = 0.25, max_nn
     = 100))
target_fpfh = compute_fpfh_feature(target_down,
    KDTreeSearchParamHybrid(radius = 0.25, max_nn
     = 100))
result_ransac =
    registration_ransac_based_on_feature_matching
    (source_down, target_down, source_fpfh,
    target_fpfh, max_correspondence_distance =
    0.075, TransformationEstimationPointToPoint(
    False), ransac_n = 4, [
    CorrespondenceCheckerBasedOnEdgeLength(0.9),
    CorrespondenceCheckerBasedOnDistance(0.075)],
     RANSACConvergenceCriteria(max_iteration =
    4000000, max_validation = 500))
```

We used profiling tools to analyze the RANSAC-based algorithm and found that its most time-consuming part is the validation of matching results. Thus, we give the user an option to specify a termination criterion via the *RANSACConvergenceCriteria* parameter. In addition, we provide a set of functions that prune false matches early, including *CorrespondenceCheckerBasedOnEdgeLength* and *CorrespondenceCheckerBasedOnDistance*.

The final part of the pairwise registration workflow is ICP refinement [2, 4], applied to the original dense point clouds:

```
result_icp = registration_icp(source, target,
    max_correspondence_distance = 0.02,
    result_ransac.transformation,
    TransformationEstimationPointToPlane())
```

Here *TransformationEstimationPointToPlane()* invokes a point-to-plane ICP algorithm. Other ICP variants are implemented as well. Intermediate and final results of the demonstrated registration procedure are shown in Figure 3.

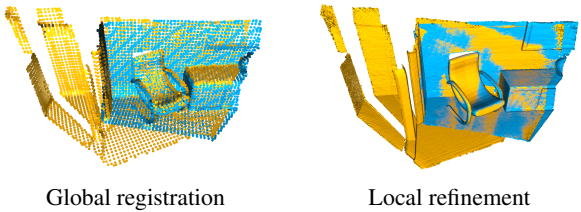

Global registration       Local refinement

Figure 3: Intermediate and final results of pairwise registration.

### 3.4. Reconstruction

A sophisticated workflow that is demonstrated in an Open3D tutorial is a complete scene reconstruction system [5, 15]. The system is implemented as a Python script that uses many algorithms implemented in Open3D. It takes an RGB-D sequence as input and proceeds through three major steps.

1. Build local geometric surfaces $\{\mathbf{P}_i\}$ (referred to as fragments) from short subsequences of the input RGB-D sequence. There are three substeps: matching pairs of RGB-D images, robust pose graph optimization, and volumetric integration.

2. Globally align the fragments to obtain fragment poses $\{\mathbf{T}_i\}$ and a camera calibration function $\mathcal{C}(\cdot)$. There are four substeps: global registration between fragment pairs, robust pose graph optimization, ICP registration, and global non-rigid alignment.

3. Integrate RGB-D images to generate a mesh model for the scene.

Figure 4 shows a reconstruction produced by this pipeline for a scene from the SceneNN dataset [10]. The visualization was also done via Open3D.

## 4. Optimization

Open3D was engineered for high performance. We have optimized the C++ backend such that Open3D implementations are generally faster than their counterparts in other 3D processing libraries. For each major function, we used profiling tools to benchmark the execution of key steps. This analysis accelerated the running time of many functions by multiplicative factors. For example, our optimized implementation of the ICP algorithm is up to 25 times faster than its counterpart in PCL [18]. Our implementation of the reconstruction pipeline of Choi et al. [5] is up to an order of magnitude faster than the original implementation released by the authors.

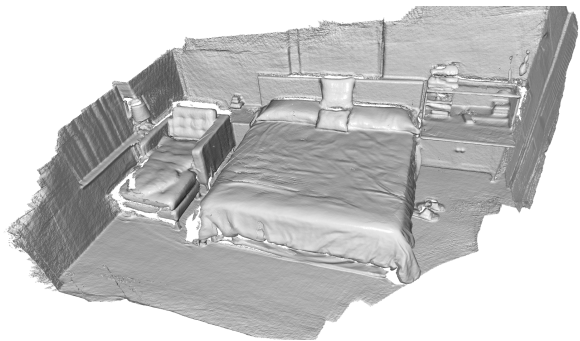

Figure 4: A scene reconstructed and visualized in Open3D.

A large number of functions are parallelized with OpenMP. Many functions, such as normal estimation, can be easily parallelized across data samples using "*#pragma omp for*" declarations. Another example of consequential parallelization can be found in our non-linear least-squares solvers. These functions optimize objectives of the form

$$L(\mathbf{x}) = \sum_i r_i^2(\mathbf{x}), \quad (1)$$

where $r_i(\mathbf{x})$ is a residual term. A step in a Gauss-Newton solver takes the current solution $\mathbf{x}^k$ and updates it to

$$\mathbf{x}^{k+1} = \mathbf{x}^k - (\mathbf{J}_\mathbf{r}^\top \mathbf{J}_\mathbf{r})^{-1}(\mathbf{J}_\mathbf{r}^\top \mathbf{r}), \quad (2)$$

where $\mathbf{J}_\mathbf{r}$ is the Jacobian matrix for the residual vector $\mathbf{r}$, both evaluated at $\mathbf{x}^k$. In the Open3D implementation, the most time-consuming part is the evaluation of $\mathbf{J}_\mathbf{r}^\top \mathbf{J}_\mathbf{r}$ and $\mathbf{J}_\mathbf{r}^\top \mathbf{r}$. These were parallelized using OpenMP reduction. A lambda function specifies the computation of $\mathbf{J}_\mathbf{r}^\top \mathbf{J}_\mathbf{r}$ and $\mathbf{J}_\mathbf{r}^\top \mathbf{r}$ for each data record. This lambda function is called in a reduction loop that sums over the matrices.

The parallelization of the Open3D backend accelerated the most time-consuming Open3D functions by a factor of 3-6 on a modern CPU.

## 5. Release

Open3D is released open-source under the permissive MIT license and is available at http://www.open3d.org. Ongoing development is coordinated via GitHub. Code changes are integrated via the following steps.

1. An issue is opened for a feature request or a bug fix.

2. A developer starts a new branch on a personal fork of Open3D, writes code, and submits a pull request when ready.
3. The code change is reviewed and discussed in the pull request. Modifications are made to address issues raised in the discussion.
4. One of the admins merges the pull request to the master branch and closes the issue.

The code review process maintains a consistent coding style and adherence to modern C++ programming guidelines. Pull requests are automatically checked by a continuous integration service.

We hope that Open3D will be useful to a broad community of developers who deal with 3D data.